\documentclass[journal]{IEEEtran}

    \usepackage{amsmath}
    \usepackage[utf8]{inputenc}
    \usepackage{graphicx}
    \usepackage{cite}
    \usepackage{tabularx}
    
    \ifCLASSINFOpdf
    \else
    \fi
    
\begin{document}
    \title{Offline handwritten mathematical symbol recognition utilising deep learning}
    
    \author{Azadeh~Nazemi\(^1\), Niloofar Tavakolian\(^2\), Donal Fitzpatrick\(^1\), Chandrika Fernando\(^3\), Ching Y. Suen\(^4\)\\
    \(^1\)School of Computing, Dublin City University,
    \(^2\)Department of Computer Science and Engineering and Information Technology, Shiraz University
    \(^3\) School of Electrical Computer and Mathematics, Curtin University
    \(^4\)Centre for Pattern Recognition and Machine Intelligence, Concordia University
            
    \thanks{ }
    }
    \maketitle
    
    
    \begin{abstract}
    This paper describes an approach for offline recognition of handwritten mathematical symbols. The process of symbol recognition in this paper includes symbol segmentation and accurate classification for over 300 classes. Many multidimensional mathematical symbols need both horizontal and vertical projection to be segmented.  However, some symbols do not permit to be projected and stop segmentation, such as the root symbol. Besides, many mathematical symbols are structurally similar, specifically in handwritten such as 0 and null. There are more than 300 Mathematical symbols. Therefore, designing an accurate classifier for more than 300 classes is required. This paper initially addresses the issue regarding segmentation using “Simple Linear Iterative Clustering”(SLIC). Experimental results indicate that the accuracy of the designed kNN classifier is 84\% for salient, 57\% Histogram of Oriented Gradient (HOG), 53\% for Linear Binary Pattern (LBP)  and finally 43\% for pixel intensity of raw image for 66 classes. 87 classes using modified LeNet represents 90\% accuracy. Finally, for 101 classes, SqueezeNet achieved 90\% accuracy using pre-trained one. 
    
    \end{abstract}
    
    
    \IEEEpeerreviewmaketitle
    \section{Introduction}
    
    \IEEEPARstart{M}{any}researchers have worked on Handwritten Mathematical Expression (HME) recognition for many years such as Anderson's effort in the '60s[1][2]
    Generally, handwritten recognition based on the applied method has two categories: Offline and Online. Many devices, such as Personal Digital Assistants(PDAs), tablets, Personal Computers( PCs), and electronic whiteboards, require online handwriting recognition. However, offline handwritten recognition is used for the scanned documents; it is less appealing than online. Therefore, at present little research is published on offline handwriting. In contrast, there are many comprehensive approaches have been applied for online HME. Competition on Recognition of Online Handwritten Mathematical Expressions (CROHME) 2016 concluded that the recognition of HMEs was still a challenge after six years of competition[3]
    The online method works well for connected strokes or cursive strokes[4], while the offline way can overcome the problem of out-of-order strokes or duplicated strokes using contextual information[5][6].
    Online HME recognition can improve by tracing some points of symbols from starting to ending as ink Markup Language (inkML) format indicates.
    Nevertheless, offline HME recognition involves two-dimensional layouts, subscript positioning, variable-sized characters, and unusual maths operators depending on the area of mathematics. 
    In addition to the variation in handwriting styles, the media such as pen and paper also may lead to more challenges[7].
    The issue regarding mathematical symbol recognition is not limited to symbol classification, and It requires interpreting the mathematical expression.  Interpretation of mathematical expression is performed based on considering its structures and mathematical area even in printed mathematical expression recognition 
    In the recent decade, deep learning is the state-of-the-art for many machine learning applications such as classification, detection, identification, and many more. Among several recognition methods, deep learning with multiple layers of nonlinear information processing is an inevitable trend. These approaches automatically learn and solve problems without using any prior knowledge[8]. 
    The results of  recent researchers studies prove that Deep neural networks improve online mathematical recognition symbols comparing previous methods like  Modified Quadratic Discriminant Functions (MQDFs) [9] with Bidirectional Long Short-term Memory (BLSTM) and Markov Random Fields (MRFs)[10][11]. However, there is still a lot of room for researchers to explore. To design and train a deep network, large scale datasets are required. This issue in many classification cases can be solved utilising augmentation. In this study, augmentation is not applicable due to the extreme similarity of some symbols. Besides HME recognition has common problems with the retrieval and recognition of complex structures [12][13].
    Based on the research for offline handwritten recognition using Neural Network(NN)[7] 
    identification and verification of off-line handwritten signatures from images is a difficult problem due to dependency of signature as bio-metrics factor  on psychological factors of the individual[14]. HME also face the same issue for recognition. This research performs symbol segmentation with Simple Linear Iterative Clustering (SLIC)[15]. Recognition uses a modified LeNet network and of pre-trained SqueezeNet. For the purpose of comparison and evaluation of the proposed method, some former methods such as Histogram of Oriented Gradient (HOG),  Linear Binary Pattern (LBP) are also implemented.
    Deep learning starts using a simple network by only tree  fully connected layers and handwritten digits of the MNIST dataset[16]. Then the MNIST dataset is upgraded to alphanumeric
    dataset (66 classes) and network is changed to LeNet. At this point, some network hyperparameters
    are modified to increase accuracy and control loss. In the next state carefully the number of classes
    is increased up to 87 by adding 21 mathematical symbols and performing supervised and limited
    augmentation. In addition, the network is modified to have three convolution layers, three Maxpooling
    as well as three fully connected layers. The last stage of using deep learning is transferring network to
    SqueeezeNet, training 101 available mathematical symbol classes applying transfer learning to
    prevent overfitting in the lack of sufficient data[17].
    The rest of this paper has been organised as follows. Section II includes related work which is the basis of this research. Section III contains the methodology followed by the experimental arech reviewed. For classifications purposes feature based  methods such  as Linear Binary Pattern(LBP)[18] and Histogram of Oriented Gradient(HOG) are studied. Besides two other approaches based on intensity and salience are considered for comparison.
    
    \subsection{Limitations of symbol recognition}
    Symbol recognition process, regardless of being online or offline, has some limitations that do not allow achieving high accuracy. These limitations are of two types as follows
        
        \begin{itemize}
      \item Type I: Symbols with different meanings and similar shapes such as:
    9q    gq	bh    GC	KR    DO	UV    QO   B8    Z2	S5    I1   Uv.
      \item  Type II: Upper and lower case symbols with same shapes such as:
    nN    fF     wW    zZ	xX    yY	cC    uU    	sS    kK    mM    pP    	vV
    \end{itemize}
        
    However, in some cases online HMEs can be properly recognised due to tracking of coordinate values from start to stop. Mis-recognition occurring in Group I have stronger negative impacts on the process of Mathematical Expression Retrieval (MER). The only avenue to accurately recognise these symbols is concept analysis performed after symbol recognition considering pre and post-located symbols.
    
    \subsection{ Segmentation}
    
    In order to obtain accurate results from mathematical symbol classification, a multi-step segmentation module is required. These steps include block segmentation, line segmentation, line classification as thress classes of text-only, embedded maths and maths-only classes thr word segmentation, and symbol segmentation. Line classification is to classify mathematical multi-dimensional lines, linear text lines,and mixed lines.
    In 2016, Nazemi’s thesis proposed Support Vector Machine (SVM) for line classification[19]. Symbol segmentation contains two sub-modules, namely, Vertical Symbol Segmentation(VSS) and Horizontal Symbol Segmentation(HSS). In complex mathematical expressions, it requires that HSS and VSS recursively be performed to extract symbols as the primary component of a mathematics. HSS and VSS are failed  for extraction some mathematical symbols such as root sign.\\

    Figure 1. shows steps of recursive HSS VSS for sample expression
    
    \begin{figure}[htp]
        \centering
      \includegraphics[scale=0.5]{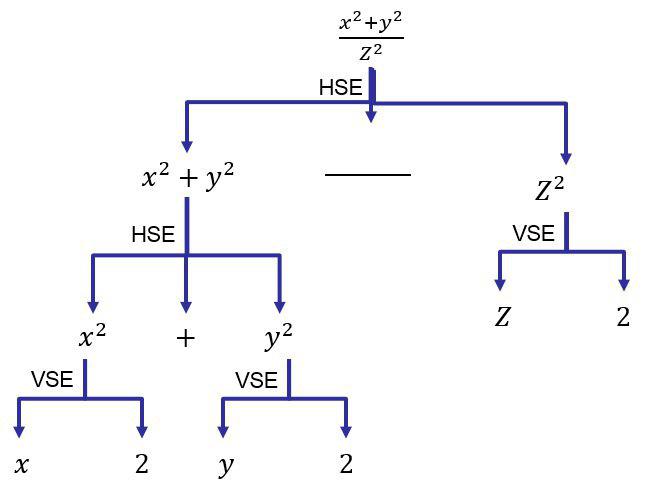}
    \caption{recursive HSS VSS for sample expression}
        \label{fig:galaxy}
    \end{figure}
     
    
    \subsection{Symbol categorisation }Symbol categorisation has been performed based on symbol’s aspect ratio. A symbol  aspect ratio is the ratio of height to width and leads to  the following Symbol categorisation 
    \begin{itemize}
    \item  if Aspect-Ratio  is greater than 1.3 then symbol is  Tall such as    $\prod$ ,$\sum$,$\int$
    
    \item  if Aspect-Ratio is between 0.58 and 1.7 then symbol is Square such as  'exist' sign ,'for all' sign and equivalent sign
    \item if Aspect-Ratio is less than 0.76 then symbol is Short such as <>- $\infty$
    \end{itemize}

    The categorisation results for 258 different symbols considering their aspect ratio were 43 “Tall”, 206 “Square” and 18 “Short”. Then each symbol was re-scaled to a predefined size as follows: \\
    the “Tall” symbol is  resized to 16x28,\\
    the “Square” symbol is converted to 28x28, and\\
    the “Short” symbol is converted to 28x16  when actual size is defined in format W x H.\\ 
    categorisation performs to implement the  more accurate classification system and reduce computational cost 
    
    \subsection{Classifiers based on HOG, LBP, Salient,  and  Intensity}
    For pixel intensity if symbol size is 28x28 then the feature vector is 784. 
    In the Salient approach, if symbol is 28x28 the saliency map is a binary image with size of 128x28[20][21]
    
     For LBP, if the symbol size is 28x28 and pixels per cell= 3x3, then the number of cells will be 9x9. Therefore, the size of feature vector is 81.  The following snippet shows  LBP  calculation:
    
    $$p_{11}\quad p_{12}\quad p_{13}$$  
       $$p_{21}\quad p_{22}\quad p_{23}$$ 
       $$p_{31}\quad p_{32}\quad\ p_{33}$$ 
    \newcommand{\blank}[1]{\hspace*{#1}}
    For LBP convert symbol image (28x28) to (3x3)\\
    pattern=''\\
    for i=1 ; i<=3; i++\\
    \blank{1cm}{for j=1 ;j<=3;j++}\\
    \blank{2cm}       if  \(p_{22}> p_{ij}\)\\
    \blank{2.5cm}        pattern=0pattern\\
    \blank{2cm}         else\\
    \blank{2.5cm}                 pattern=1pattern\
              
    For HOG if the symbol size is 28x28 and pixels per cell= 9x9, then the number of cells will be 3x3.Then value of orientation and cell per blocks determine the size of feature vector. Supposing the orientation=9 and cell per block=4, then the size of feature vector is 9x4=36. 
    In the MNIST
    dataset of zero to nine handwritten digits, the plainest method for designing a classifier is converting each symbol image to the vector of intensity and using k-nearest neighbour. It has an accuracy of 96.88\% . Although this method has a reasonable accuracy for a small number of classes, it canbe expected to keep the accuracy high when the number of classes is more than 300 due to the limitations of symbol recognition. To obtain most discriminators features, HOG or LBP can be used. During implementation of feature extraction in this classifier, a restriction was observed due to disability to extract such features from symbols. To address this issue, a morphology technique was performed to change the boldness of symbols[22] Figure 2 illustrates some symbols before and after morphology dilation.
    \begin{figure}[htp]
        \centering
      \includegraphics[width=6cm]{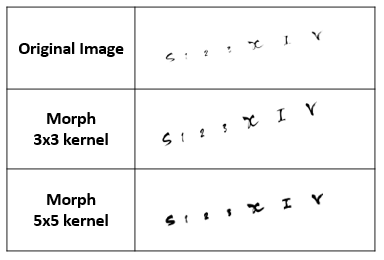}
    \caption{Applying morphology technique to change the boldness of symbols}
        \label{fig:galaxy}
    \end{figure}
    
    In order to implement classifiers, this research assembled a dataset of alpha numerical handwritten symbols (66 symbols) together with 21
    mathematical symbols belonging to three different categories, namely,‘short’, ‘tall’ and ‘square’.The results pixel intensity,salient, features extracted using HOG and features extracted using LBP are presented for comparison.The results indicate that the accuracy using kNN classifier is 84\% for salient, 57\% for HOG, 53\% for LBP and finally 43\% for pixel intensity for the 87 classes of mathematical handwritten
    symbols.
    The following figures (Figures 3,4) illustrate the results of t-distributed Stochastic Neighbours Embedding \(t_{SNE}\) scatter to visualise high-level representations obtained with intensity (raw image), and salience \(t_{SNE}\)illustrations confirmed the obtained accuracy for two methods.
    
    \begin{figure}[htp]
        \centering
      \includegraphics[width=8.5cm]{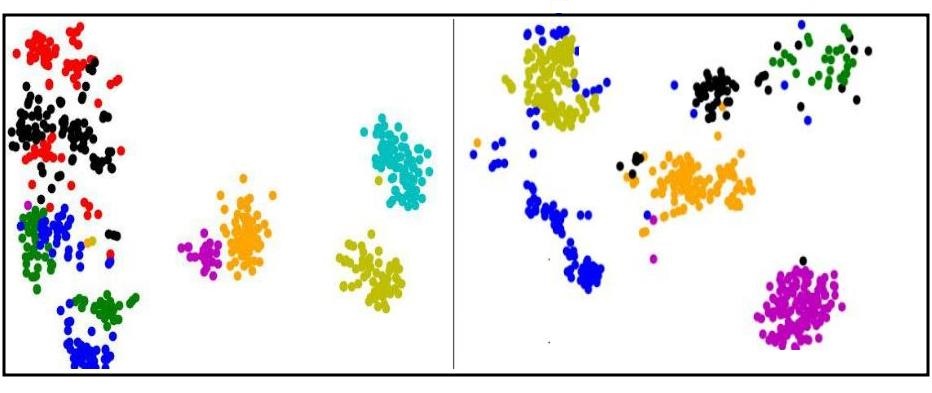}
    \caption{Salient visualisation for 10 random classes}
        \label{fig:galaxy}
    \end{figure}
    \begin{figure}[http]
      \centering
      \includegraphics[width=8.5cm]{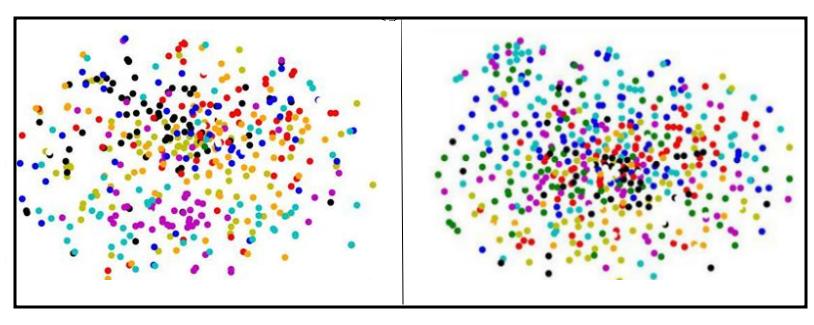}
    \caption{Pixel intensity  visualisation  for 10 random classes}
        \label{fig:galaxy}
    \end{figure}
    Fig 3 illustrates the inner class similarity of menders is more than their similarities to other classes (for saliency map).
    Fig 4 illustrates How the similarity of members of two different classes  are more than the members inner class similarity(for intensity vectors)  
    \subsection{Used dataset}
    The dataset used in this study was made of MNIST[10], 6000 images from inkML of the CROHME dataset converted to PNG by the developed converter in this research, 2000 images from the set of Handwritten Digit Images published by Computer Vision Group of the University of São Paulo and 1000 handwritten character symbols from the Chars74K dataset
    
    \section{Proposed method}
    This section initially presents the proposed method in this study for symbol segmentation followed by a description of all steps of utilising and modifying designed convolutional network for classification[23][24] over 100 classes in detail.
    
    \subsection{Segmentation with SLIC}
    Simple Linear Iterative Clustering(SLIC) Superpixel is a group of connected pixels with similar colours or gray levels. Superpixel segmentation divides an image into hundreds of non-overlapping superpixels. For the first time, in 2003,  and Malik used superpixels to do image segmentation. Superpixels provide access to meaningful regions for accurate feature extraction and reduce the input entities for further processing.
    Superpixel segmentation is utilised for semantic segmentation, visual tracking, image classification, and many computer vision applications. SLIC is an approach that utilises clustering for superpixel segmentation by combining five-dimensional colours and image plane space. It is a method to replace multi-step pre-processing for segmentation. SLIC can solely perform the role of all of the steps stated above to obtain segmented symbols.  The SLIC method can address the issues regarding segmentation of some symbols such as binding root as Figure 5 illustrates.
    \begin{figure}[htp]
        \centering
      \includegraphics[width=4cm]{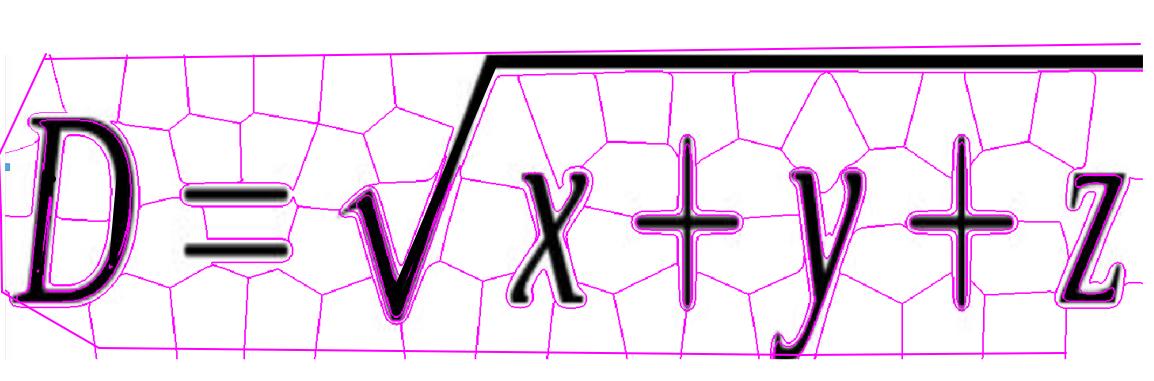}
    \caption{SLIC segmentation}
        \label{fig:galaxy}
    \end{figure}
    \subsection{Classification}
    In this subsection, two different architectures are described as the approaches for mathematical handwritten symbol recognition.
    Data augmentation means increasing the size of training dataset by changing color channels, adding noise, applying random regional sharpening filters, and adding mean-images based on clustering techniques efficiently. Most recent augmentations are centered on using the Generative Adversarial Network (GAN) model, sometimes swapping the generator network with a genetic algorithm. However, to perform augmentation on a dataset of handwritten mathematical symbols, it must be considered that augmentation does not change the symbol meaning, for example,\(\cup\) when is horizontally flipped, it is converted to \(\cap\). Therefore, augmentation techniques cannot be run on all symbols. In the result, this research faces lack of sufficient data for deep learning. On the other hand, due to regularising effect of augmentation, by increasing the usage of augmented samples, the possibility of network under fitting increases. Thus, data augmentation is not always efficient and is not an ideal approach to address the lack of sufficient training sample for deep convolutional network. Figure 6 illustrates applying shift augmentation techniques which is almost safe for handwritten mathematical symbol recognition.
    \begin{figure}[ht]
        \centering
      \includegraphics[width=9cm]{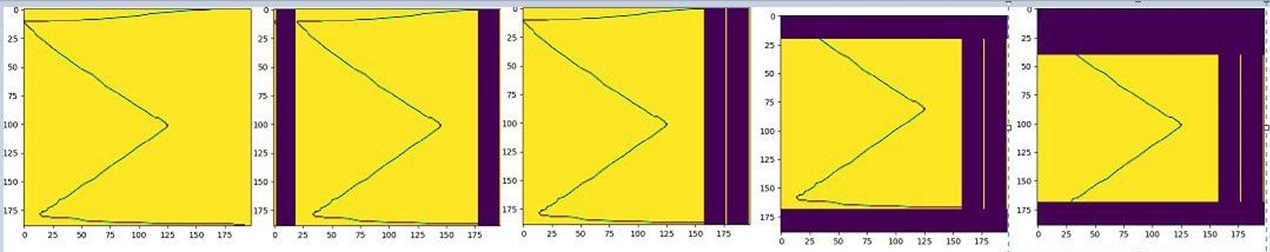}
    \caption{Applying shift augmentation techniques.}
        \label{fig:galaxy}
    \end{figure}
    
    Considering the above mentioned obstacles in the design of a deep network to classify handwritten mathematical symbols, this research started with networking from a very simple network with only 10 classes using MNIST large scale dataset for digits only and having sufficient samples. Since training dataset was enough and only 10 classes need to recognise in this case, even the network with only three fully connected layers (2 RelU and 1 Softmax) and without any convolutional layer, provided at least 89\% accuracy. The empirical results of the research demonstrated that the transmission of the network from a very simple network to LeNet with seven layers including convolutional layers, even with changing MNIST with 10 classes to alphanumeric dataset still can maintain high accuracy. Alphanumeric dataset contains the numbers 0 - 9, the letters A - Z (both uppercase and lowercase), and some common symbols such as sharp sign,ampere sign, @, and *.  Totally, there are 66 alphanumeric case sensitive symbols.
    After segmentation, the next major objective for this study is symbol classification. As long as the number of classes is less than 40,  symbol image saliency  map followed by kNN classifier, have reasonable classification results. Moreover, modification of some conevolutional networks with a few layers such as LeNet, makes the system efficient for less than 40 classes.
    The most prominent challenge for classification of the handwritten mathematical symbols is the real world nature of the training samples. In some cases, the extracted descriptor from the last layer of the network proves inner class discrimination more than the outer class. Another issue to design a neural network to meet the requirement for specific task classification is the lack of sufficient training samples and imbalance training data sets with different number of samples in each class. To address the first issue, there are two avenues, one is augmentation and the other is Transfer Learning (TL)[25] which means  the weights of the pretained networks with the large scale training data. Making the decision to use one of these two approaches fairly depends upon the nature of training sample.
    To find out most appropriate network architecture to classify handwritten mathematical symbols, this research initially visualise the output of the last convolution layer for some networks such as VGG16,  LeNet, SqueezeNet applied on some printed symbols. The results indicated that the VGG networks were not suitable for this purpose. However, LeNet and SqueezNet had better functionality.  Experimental result of this comparison is presented in Figure 7 .
    \begin{figure}[ht]
        \centering
      \includegraphics[scale=0.75]{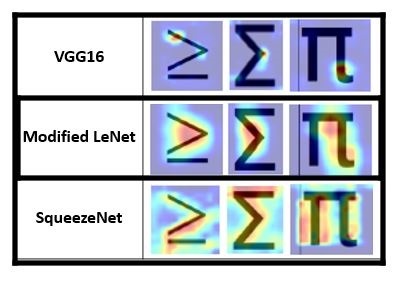}
    \caption{VGG,LeNet, SqueezeNet comparison.}
        \label{fig:galaxy}
    \end{figure}
        Gradient-weighted Class Activation Mapping (Grad-CAM)  technique used for this visualisation and includes following steps.
      \begin{itemize}
    
      \item    Picking final convolutional feature map 
     \item   Considering a sample image  
    \item     Predicting  its class by the network
    \item    Calculating the gradient of the class noting 
    width  height  and colour channels
    \item  Updating  the weights of each channel  in the sample feature map using class gradient
    \end{itemize}[26]

    \subsection{ LeNet}
    The proposed network architecture, LeNet, for deep learning for classification of alphanumeric symbols originally used tangent hyperbolic (tanH) as the activation function. Considering the number of classes, this research used rectified linear unit (RelU) to improve classification accuracy.
    In the following equation: O=Output, W=Input, F=Filter Size, P=Padding, and S=Stride
    \begin{equation}
    O=\frac{(W-F+2P)}{S}+1
    \end{equation}
    The starting image size was 32x32 with filter size 5x5,  Stride 2 and padding was equal to zero. Equation 1 was used to obtain output size by initial size, filter, stride and padding. Based on Equation 1, the size of the next layer was calculated as 28x28. In the next setup, for the convolutional layer, the filter size is 2x2, the padding is zero and stride was two which lead to a size of 14x14.
    Generally, the LeNet architecture contains two sets of convolutional layers, activation, and pooling, followed by two fully connected activation and end up to classifier function such as Softmax. These sequential network included input, conv1, RelU, pool1, conv2, RelU, pool2, FC, RelU, FC, Softmax, and Output respectively. This statement is further illustrated in  Figure 8  and Table I  in detail.
    \begin{figure}[ht]
        \centering
      \includegraphics[width=8cm]{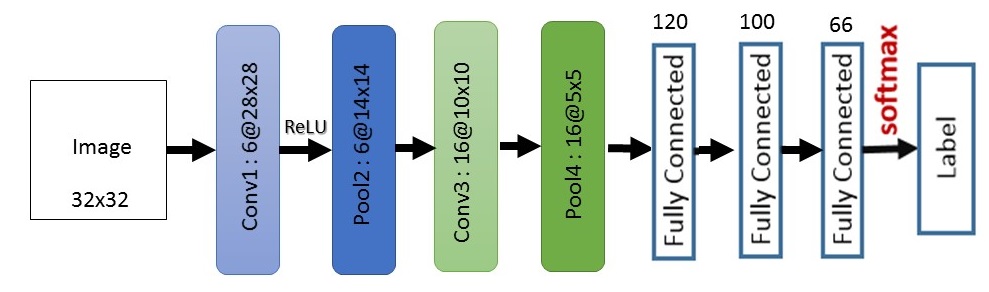}
    \caption{The original LeNet used for classifying 66 classes.}
        \label{fig:galaxy}
    \end{figure}
    \begin {table}
    \caption{Seven layers of LeNet}
    \label{table_2}
    \centering
    
    
    \begin{tabular}{|c|c|}
    \hline
     Layers & Description \\
    \hline
    Input & Size=32x32\\
    \hline
    Convolutional 1 & I=32x32\hspace{0.1cm}\vline\hspace{0.1cm} P=0\hspace{0.1cm}\vline\hspace{0.1cm} F=5\hspace{0.1cm}\vline\hspace{0.1cm} O=28x28\hspace{0.1cm}\vline\hspace{0.1cm} Feature=6\\
   \hline
    Activation  & RelU / tanH\\
    \hline
    Pooling 1 & I=28x28\vline S=2\hspace{0.1cm}\vline\hspace{0.1cm}P=0\hspace{0.1cm}\vline\hspace{0.1cm}F=2\hspace{0.1cm}\vline\hspace{0.1cm} O=14x14\\
    \hline
    Convolutional 2 & I=14x14\hspace{0.1cm}\vline\hspace{0.1cm}S=1\hspace{0.1cm}\vline\hspace{0.1cm} P=0\hspace{0.1cm}\vline\hspace{0.1cm} F=5\hspace{0.1cm}\vline\hspace{0.1cm} O=10x10 \hspace{0.1cm}\vline\hspace{0.1cm} Feature=6\\
    \hline
    Activation& RelU / tanH\\
    \hline    
    Pooling 2 & I=12x12\hspace{0.1cm}\vline\hspace{0.1cm} S=2\hspace{0.1cm}\vline\hspace{0.1cm}
     P=0\hspace{0.1cm}\vline\hspace{0.1cm}  F=2\hspace{0.1cm}\vline\hspace{0.1cm} O=5x5\\
    \hline
    Fully connected &  Input=400\\
    \hline 
    Fully connected & Input=200 \\
    \hline
    Loss function &  Sigmod/Softmax \\
    \hline
    Output & Size= 66 class\\
    \hline
    \end{tabular}
    \end{table}

    Figure 9 and  Table II both describe three layers insertion to LeNet(modified LeNet) for increasing accuracy.
    \begin{figure}[ht]
        \centering
      \includegraphics[scale=0.25]{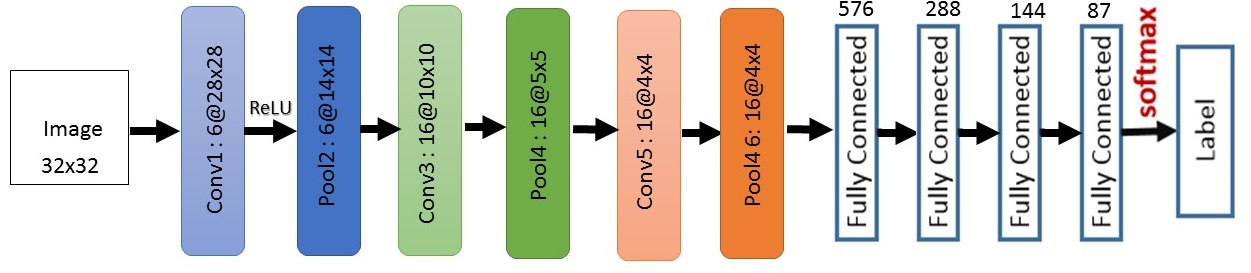}
    \caption{The modified LeNet for classifying 87 classes}
        \label{fig:galaxy}
        
    \end{figure}
    
    Table III compares different performance based on different hyper parameters in terms of  Accuracy.                               
    
    \begin{table}
    \caption{Modified ten layers of LeNet}
    \label{table_3}
    \centering
   \begin{tabular}{|c|c|}
    \hline
    Layers & Description \\
    \hline
    Input&	Size= 32x32\\
    \hline
    Convolution 1&	I=32x32\hspace{0.1cm}\vline\hspace{0.1cm} S=1\hspace{0.1cm}\vline\hspace{0.1cm} P=2\hspace{0.1cm}\vline\hspace{0.1cm} F=9\hspace{0.1cm}\vline\hspace{0.1cm} O=28x28\hspace{0.1cm}\vline\hspace{0.1cm} Feature=6\\
    \hline
    Activation &RelU / tanH\\\hline
    Pooling 1 & I=28x28\hspace{0.1cm}\vline\hspace{0.1cm} S=2\hspace{0.1cm}\vline\hspace{0.1cm} P=2\hspace{0.1cm}\vline\hspace{0.1cm} F=2\hspace{0.1cm}\vline\hspace{0.1cm} O=16x16\\
    \hline
    Convolution 2& I=16x16\hspace{0.1cm}\vline\hspace{0.1cm} S=1\hspace{0.1cm}\vline\hspace{0.1cm} P=2\hspace{0.1cm}\vline\hspace{0.1cm} F=9\hspace{0.1cm}\vline\hspace{0.1cm} O=12x12\hspace{0.1cm}\vline\hspace{0.1cm} Feature=16\\    \hline
    Activation &	RelU / tanH\\
    \hline
    Pooling 2&	I=12x12\hspace{0.1cm}\vline\hspace{0.1cm} S =2\hspace{0.1cm}\vline\hspace{0.1cm}P=2\hspace{0.1cm}\vline\hspace{0.1cm} F=2\hspace{0.1cm}\vline\hspace{0.1cm} O=8x8\\
    \hline
    Convolution 3& I=8x8\hspace{0.1cm}\vline\hspace{0.1cm} S=1\hspace{0.1cm}\vline\hspace{0.1cm} P=2\hspace{0.1cm}\vline\hspace{0.1cm} F=5\hspace{0.1cm}\vline\hspace{0.1cm} O=8x8\hspace{0.1cm}\vline\hspace{0.1cm} Feature=16\\
    \hline
    Activation &	RelU / tanH\\
    \hline
    Pooling 3&I=8x8\hspace{0.1cm}\vline\hspace{0.1cm} S=2\hspace{0.1cm}\vline\hspace{0.1cm}P=2\hspace{0.1cm}\vline\hspace{0.1cm} F=2\hspace{0.1cm}\vline\hspace{0.1cm} O=6x6\\
    \hline
    Fully connected&	Input=576\\
    \hline
    Fully connected& Input=288\\
    \hline
    Fully connected& Input=144\\
    \hline
    Loss function &    Softmax\\
    \hline
    Output&  size=87 Classes\\
    \hline
    \end{tabular}
    \end{table}
    
    \begin{table}
    \caption{LeNet/Modified and Hyper parameters}
    \label{table_4}
    \centering
    \begin{tabular}{|c|c|c|c|c|c|}
    \hline
    F(Activation)&F(Loss)& F(Optimisation)& LR& Accuracy&Classes\\
    \hline
    tanH& Sigmoid&SGD&0.01&88\% & 66\\
    \hline
    RelU& Softmax&SGD&0.01&80\% & 66\\
    \hline
    RelU& Softmax&ADAM&0.00001&80\% & 66\\
    \hline
    tanH& Sigmoidx&ADAM&0.00001&90\% & 66\\
    \hline
    RelU& Sigmoid&ADAM&0.00001&90\% & 66\\
    \hline
    RelU& Sofmax&ADAM&0.00001&90\% & 87\\
    \hline

    \end{tabular}
    \end{table}
    
    Changing the parameters such as learning rate(0.01-0.00001), optimization function (ADAM and SGD), activation functions (RelU, tanH), loss functions (softmax and sigmoid) resulted in different recognition rates upto 90\%.Then, gradually the number of classes was increased from 66 to 87 with the inclusion of alpha numeric symbols and 7 symbols of each of the three categories, namely, ‘tall’, ‘square’ and ‘short’ .The same set of changes resulted in lower recognition rates.In conclusion, changing parameters for 66 classes achieved 90\% accuracy, but further increase of the number of classes caused a drop in accuracy. Therefore, in the next stage, LeNet structure was modified by adding one a fully connected,a convolutional layer and a pooling as illustrated in Table II. Parameters such as the size of convolution were revised and maxpooling from two series was changed to three series. Additionally, extra fully connected layer with loss function RelU was added before the last layer of softmax. All padding from zero size was changed to two. All these modifications lead to a change of filter size as indicated in Table II. Figures 8 and 9 illustrate the layer architectures of original LeNet for 66 classes and modified LeNet for 87 classes produced in this study. 
    Table III indicates all changes in Hyper parameters and related results in LeNet and  Modified LeNet with 10 layers.

    \subsection{ SqueezeNet}
      
    This subsection describes SqueezeNet, an alternative architecture for the network used in order to classify more than 100 classes up to 300.
    SqueezeNet is a small CNN architecture that achieves AlexNet-level accuracy on 1000 classes of ImageNet with 50x fewer parameters. SqueezeNet replaces 3x3 filters in AlexNet with 1x1 filters. The convolutional layer with 1x1 filters is called squeeze layer. These squeeze layers feed expand layers which contain convolutional layers with a mix of 1x1 and 3x3 filters. Fire module, the basic building block of SqueezeNet, is made of squeeze layer and expand layer.
    The fire module has two functions to perform:\\
        1. Reducing the total number of parameters and making a smaller model \\
        2. Making extra learning step to be embedded within the transformations between several of “fire modules” in the lack of fully-connected layers.\\
        The fully connected layers learn the relationships between the earlier higher level features of a CNN and the classes the network is trying to identify. That is, the fully connected layers are the ones that learn the details of symbols in most networks such as VGG.
    SqueezeNet reduces the number of parameters by decreasing the S with later convolutional layers. It creates a larger activation/feature map later in the network and increases classification accuracy. SqueezeNet is in contrast to VGG that activation maps get smaller as get closer to the end of a network.  He and Sun (2014) applied a delayed down sampling to gain higher classification accuracy[27]
    SqueezeNet is a CNN with 18 layers.It is trained on  the ImageNet database to classify 1000 classes.The pre-trained network is a version of SqueezNet which has been trained over more than one million images on ImageNet dataset for 1000 object classes. Since a pre-trained network has appropriate weights, then the technique is implemented by freezing some layers. In the next step, a network using a pre-trained model has been fine-tuned to be converged optimally based on 101 classes of mathematical symbols.
        \section
        { Experimental results}

           To select most appropriate parameters for model in network, objective function must be monitored to observe  how  in will be affected by changing parameters.This observation  support system to find proper direction  for iteration to reduce error.
        This section initially presents different plots to illustrate how various parameters affect.
        Sigmoid value is between (0 to 1). Thus it suitable to predict the probability as an output.On the other hand  Sigmoid as loss function and tangent hyperbolic(tanH) as activation function in networks with many layers based on  vanishing gradient issues are not efficient.

    Figure 10  shows the LeNet performance with the activation function Relu and the accuracy 80\% with loss function softmax, and Figure 11  shows the LeNet performance with the activation function tanH the accuracy 88\% with loss function sigmoid.
     
    \begin{figure}[ht]
        \centering
      \includegraphics[width=8cm]{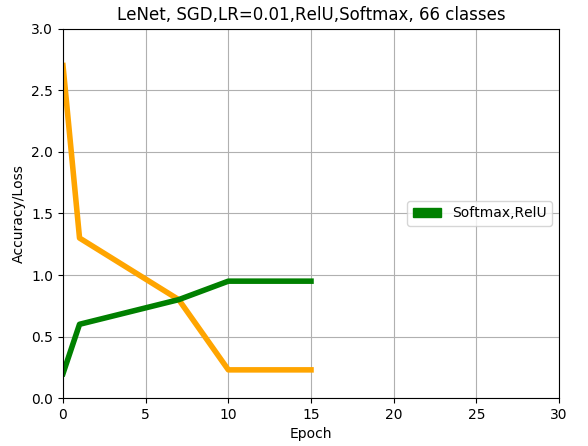}
    \caption{ LeNet - 66 classes,LR=0.01,SGD,RelU,Softmax}
    \label{fig:galaxy}
    \end{figure}
    \begin{figure}[ht]
        \centering
      \includegraphics[width=8cm]{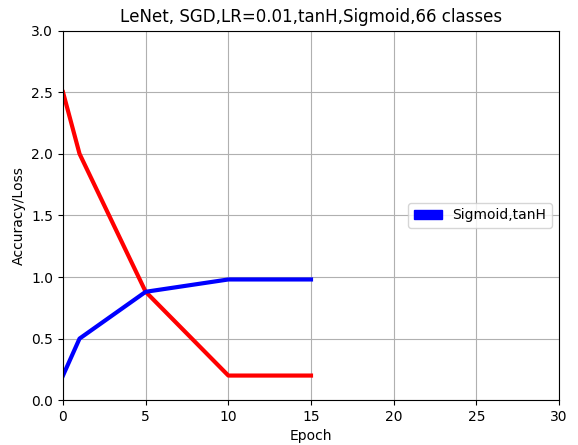}
    \caption{ LeNet-66 classes,LR=0.01,SGD,tanH,Sigmoid}
    \label{fig:galaxy}
    \end{figure}
    Since ADAM learns faster than SGD, the starting learning rate for ADAM must be much less than that of SGD[28]. The learning rate is a critical concept and must be selected carefully. If it is a short learning process, then it does not improve, and if it is a long one, then the learning process cannot converge.[29] 
      learning rate of 0.00001 and using ADAM compares ADAM and SGD as optimization
    functions for learning rate 10-5 and 10-2 respectively on LeNet. Since ADAM learns faster than SGD, starting learning
    rate for ADAM must be much less than that of SGD. Learning
    rate is a critical concept and must be selected carefully. If it is  a short learning process then it does not improve and if it is a long one, then the learning process cannot converge.

    Figure 12 shows the LeNet performance with the activation function Relu and the accuracy 80\% with loss function softmax, and Figure 13 shows the LeNet performance with the activation function tanH the accuracy 90\% with loss function sigmoid.
     
    \begin{figure}[ht]
        \centering
      \includegraphics[width=8cm]{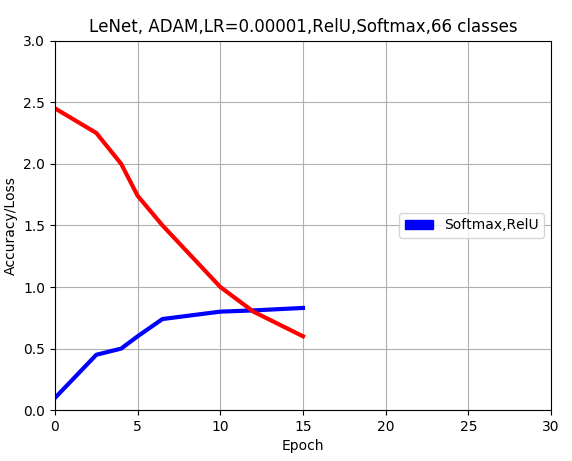}
    \caption{ LeNet - 66 classes,LR=0.00001,ADAM,RelU,Softmax}
    \label{fig:galaxy}
    \end{figure}
    \begin{figure}[ht]
        \centering
      \includegraphics[width=8cm]{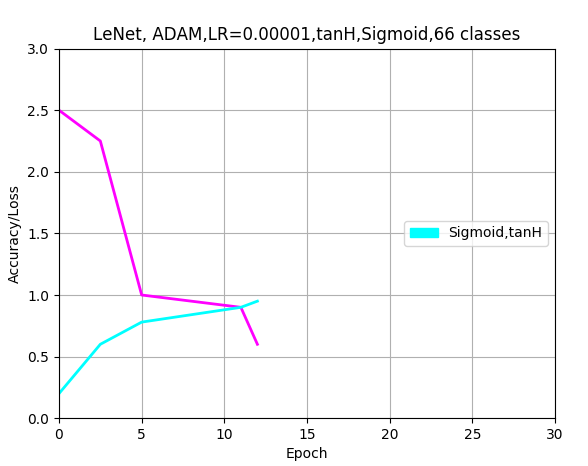}
    \caption{ LeNet-66 classes,LR=0.00001,ADAM,tanH,Sigmoid}
    \label{fig:galaxy}
    \end{figure}
    Figure 14 shows modified  LeNet run over 78 classes. Due to increase the number of layers, Sigmoid and tanH are not the best  options so Softmax and RelU are  replaced and ADAM with learning rate =0.00001 has 85\% accuracy.
    
    \begin{figure}[ht]
        \centering
      \includegraphics[width=8cm]{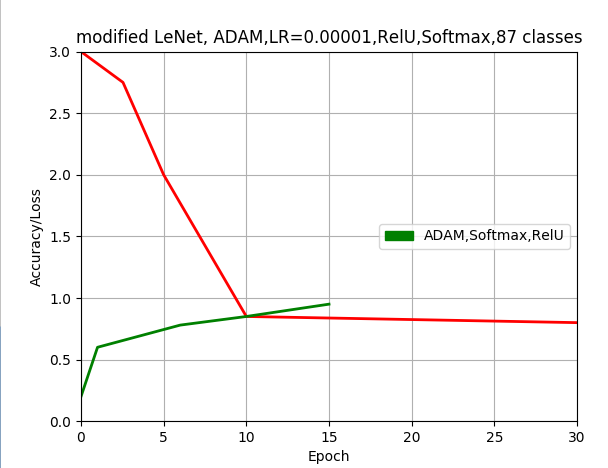}
     \caption{ Modified  LeNet-76 classes classes,LR=0.00001}    \label{fig:galaxy}
    \end{figure}
    Figure 15 shows the performance of SqueezeNet using pre-trained one over 101 classes. The pre-trained network is a version of SqueezNet which has been trained over more than one million images on ImageNet dataset for 1000 object classes[30].
    
    \begin{figure}[ht]
        \centering
      \includegraphics[width=8cm]{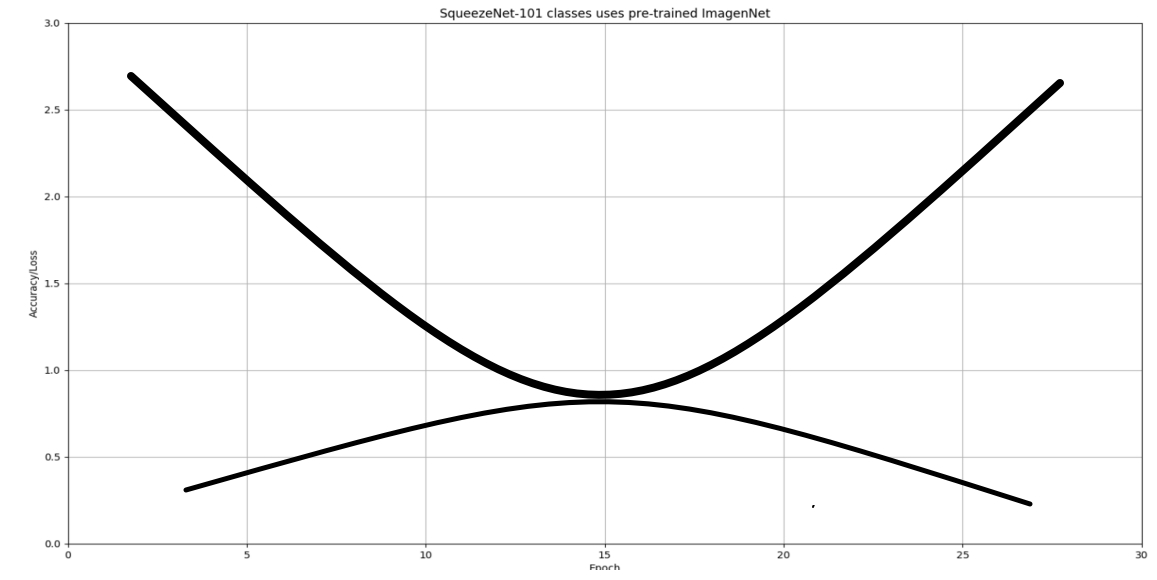}
    \caption{ SqueezeNet-101 classes classes}    \label{fig:galaxy}
    \end{figure}
    \section{Conclusion}
    The dataset used in this study was made of MNIST  6000 images from inkML of the CROHME dataset converted to PNG by the developed application in this research, 2000 images from the set of Handwritten Digit Images published by Computer Vision Group of the University of São Paulo
    In this study, two different types of classification methods, were used for offline handwritten mathematical symbol recognition. Intensity of raw image as well as extracted features using HOG, LBP, and salient maps were used to see the performance with regard to recognition. The Neural Network methods used for this research were LeNet and SqueezeNet. The number of classes was increased gradually 66 to 101. The final experimental result is 90\% using pre-trained SqueezNet by transfer learning. The future work of the research is aimed at increasing the number of classes to 300. 
    .
    \ifCLASSOPTIONcaptionsoff
      \newpage
    \fi

    \end{document}